\pdfoutput=1

\documentclass[11pt]{article}

\usepackage[preprint]{acl}

\usepackage{times}
\usepackage{latexsym}

\usepackage[T1]{fontenc}

\usepackage[utf8]{inputenc}

\usepackage{microtype}

\usepackage{inconsolata}

\usepackage{graphicx}

\usepackage{amsmath} 
\usepackage{dsfont}

\usepackage{colortbl}
\usepackage{xcolor}
\usepackage{pgfplotstable}
\usepackage{booktabs}

\usepackage{pgfplots}
\usepackage{etoolbox}

\title{MindShift: Analyzing Language Models' Reactions to Psychological Prompts}

\author{
 \textbf{Anton Vasiliuk
 \textsuperscript{1}},
 \textbf{Irina Abdullaeva\textsuperscript{1,3}},
\\
 \textbf{Polina Druzhinina\textsuperscript{1,2}},
 \textbf{Anton Razzhigaev\textsuperscript{1}},
 \textbf{Andrey Kuznetsov\textsuperscript{1,3}},
\\
 \textsuperscript{1}FusionBrain Lab, 
 \textsuperscript{2} Applied AI Institute, 
 \\
\textsuperscript{3} 
 Innopolis University,
\\
 \small{
   \textbf{Correspondence:} \href{abdullaeva@fusionbrain.com}{abdullaeva@fusionbrain.com}
 }
}

\begin{document}
\maketitle
\begin{abstract}
  Large language models (LLMs) hold the potential to absorb and reflect personality traits and attitudes specified by users. In our study, we investigated this potential using robust psychometric measures. We adapted the most studied test in psychological literature, namely Minnesota Multiphasic Personality Inventory (MMPI) and examined LLMs’ behavior to identify traits. To asses the sensitivity of LLMs' prompts and psychological biases we created personality-oriented prompts, crafting a detailed set of personas that vary in trait intensity. This enables us to measure how well LLMs follow these roles. Our study introduces \textbf{MindShift}, a benchmark for evaluating LLMs’ psychological adaptability. The results highlight a consistent improvement in LLMs’ role perception, attributed to advancements in training datasets and alignment techniques. Additionally, we observe significant differences in responses to psychometric assessments across different model types and families, suggesting variability in their ability to emulate human-like personality traits. \textbf{MindShift} prompts and code for LLM evaluation are open-sourced\footnote{\url{https://github.com/FusionBrainLab/MindShift}}.
\end{abstract}

\section{Introduction}

Recent advances in large language models (LLMs) have dramatically improved their ability to generate human-like text, making them essential tools for tasks ranging from personal assistance to complex dialogue systems. As LLMs increasingly interact with humans, understanding their behavioral and psychological characteristics—emerging from their training data—has become a critical research challenge. While LLMs lack intrinsic psychological traits, their ability to mimic human behavior and respond to role-based prompts raises important questions about how they encode and reflect patterns of thought, emotion, and behavior. These questions are relevant not only to psychology but also to knowledge discovery, which aims to uncover hidden patterns in data.

Psychometrics, the science of psychological measurement, offers a robust framework for evaluating personality and behavior through structured tests \cite{rust2014modern, furr2021psychometrics}. Recent studies have used psychometric tools to show that LLMs can exhibit human-like traits and values \cite{huang2023humanity, jiang2023evaluating}. However, these efforts have mostly examined surface-level traits, leaving deeper psychopathological dimensions unexplored. To address this, we adapt the Minnesota Multiphasic Personality Inventory (MMPI-2) \cite{mmpi}, a widely used psychometric tool, to analyze LLM behavior. This adaptation not only probes the psychological dimensions of model outputs but also applies data mining methods to uncover latent behavioral patterns.

In this paper, we bridge psychometrics and knowledge discovery through a data-driven framework for evaluating LLMs using psychological prompts. Our approach involves two key contributions: (1) the adaptation of the MMPI-2 to assess psychopathological traits in LLMs, and (2) the development of a structured prompting framework that introduces psychologically oriented biases to measure how LLMs interpret and respond to personality-driven instructions. By systematically varying the degree of psychological instruction, we create a benchmark, called MindShift, that evaluates the susceptibility of LLMs to psychological cues. This benchmark not only provides a tool for assessing LLM behavior but also uncovers patterns in how models evolve with advancements in training data and alignment techniques.

Our analysis advances data mining techniques that leverage language models by offering significant insights into the behavioral patterns of LLMs. Our findings reveal several key insights:

\begin{enumerate}
 \item Language models form distinct clusters based on their responses to the test, indicating variations in performance that depend on the model family and its fine-tuning.
  \item LLMs exhibit a consistent improvement in role perception over time, correlating with advancements in dataset quality and alignment techniques.
  \item Certain test scales, such as Defensiveness, Family Problems and Bizarre Mentation exhibit a strong correlation with safety and reasoning benchmarks.
  \item The length of LLM responses strongly correlates with the Defensiveness scale, while response variability exhibits a strong correlation with the Depression scale, suggesting a measurable link between response patterns and psychological traits.
\end{enumerate}


By situating our work within the framework of knowledge discovery, we demonstrate how psychometric tools can uncover meaningful patterns in LLM behavior. Our approach advances understanding of LLM psychology and offers a novel methodology for aligning AI systems with human values in psychologically sensitive contexts.

\section{Background}

Psychometrics in LLMs has been applied in various ways: studying AI personality \cite{huang2024humanity, lee2024trait}, assessing personality on text stylistics \cite{wang2023incharacter}, ensuring AI safety \cite{zhang2024better} and shaping individual predictions \cite{jiang2023personallm} with desired outcomes. Researchers have utilized psychological tests to assess these aspects, providing insights into how LLM traits influence their predictions and interactions. For example, studies have explored whether psychological tests can help identify undesirable antisocial behaviors in LLMs \cite{li2022does} \cite{reuben2024assessment}, contributing to AI safety. Other research has shown that inducing anxiety in large language models increases exploration and bias \cite{coda2023inducing}, highlighting the complex interplay between emotional states and model behavior.  Another promising direction is character training,  which offers an alternative to reinforcement learning from human feedback (RLHF) by steering a model to a desired personality \cite{claude}. Adding another psychological test, such as the MMPI, enhances research across all these fields for evaluating and applying LLMs. 

To apply psychological tests correctly to LLMs, it is crucial to ensure both reliability and validity. Reliability refers to the consistency of responses, while validity ensures that a test measures the intended properties \cite{huang2023revisiting}. Several studies have highlighted challenges in applying psychological tests to LLMs, such as sensitivity to different prompts, inconsistent responses, and differences from human cognition that may affect the reliability of these tests \cite{dorner2023personality, gupta2023investigating, song2023have, shu2023you}. While prior research has demonstrated the reliability and validity of psychometric evaluations for large-scale models (e.g., GPT-3.5-Turbo, GPT-4, and PaLM-62B) \cite{huang2023revisiting, safdari2023personality}, the applicability of these findings to some smaller LLMs (e.g., 7B-9B parameter models) has been questioned due to differences in model scale and architecture.  However, LLMs are continuously improving, particularly smaller models, and the validity of any psychometric assessment depends on the specific test being used, making it necessary to evaluate each case individually. Nevertheless, these challenges have not precluded other researchers from successfully utilizing psychological assessments to evaluate the \textit{language models personality}\cite{john1991big} and benchmark their emotional behavior\cite{huang2024humanity}.

Additionally, our methodology addresses these concerns by avoiding direct questions about psychological traits, thereby reducing the influence of Sycophancy Bias — a common issue in LLM evaluations. Furthermore, we conducted additional studies to validate the reliability of our approach when applied to smaller models, many of which have demonstrated superior performance compared to larger models like PaLM-62B across various benchmarks. This suggests that smaller, more recent models can indeed exhibit reliable and valid psychometric properties when evaluated using appropriate methodologies.

Moreover, as validity implies reliability \cite{huang2023revisiting}, ensuring a test's validity is sufficient for meaningful evaluation. This is frequently achieved through personality prompting \cite{safdari2023personality, jiang2024evaluating, huang2023revisiting, lu2023illuminating}, a method where a specific personality is shaped within the model to ensure the test accurately measures the desired traits. In our work, we utilize a similar personality prompting methodology to validate the application of the Minnesota Multiphasic Personality Inventory (MMPI) to LLMs, demonstrating its effectiveness across a range of model sizes.

Most current work exploits self-report test like Big Five Inventory (BFI) \cite{john1991big}, which consist of few scales, where a subject should rate its personality within a set of questions. Many such tests are covered by the comprehensive benchmark, PsychoBench \cite{huang2024humanity}, which consists of 13 tests, as well as by TRAIT \cite{lee2024trait}, which is enhanced with a knowledge graph to test personality in a variety of real-life scenarios. However, self-assessment is subject to biases \cite{wang2023incharacter}, as, for example, subjects may report their desired personality rather than their actual one. One solution to this problem is using Role Playing Agents \cite{wang2023incharacter}, where authors ask open-ended questions and use another LLM to assess personality based on the output. However, this method is more costly and complex, requiring longer predictions and another inference to assess personality.

In contrast, the studied MMPI test, does not have explicit questions, making their purpose unclear to the interviewee and thereby reducing individual bias. Additionally, the MMPI includes embedded validity scales, which other tests do not have, further enhancing the test's validity. Finally, MMPI's unique set of well-studied scales, different from the other tests, makes it a valuable tool for more accurate and reliable assessment of LLM traits and behaviors.
\section{Methodology}

\subsection{Psychological test adaptation}

To reliably validate LLMs' implicit understanding of psychological personality traits, it is crucial to adapt psychological scales and tailor questions specifically for language models. When LLMs are asked direct questions about inner worlds, morality, and behavioral patterns, they may exhibit biased behaviors due to extensive alignment tuning, which can result in inconsistent and unrepresentative questionnaire outcomes.

To assess the susceptibility of LLMs to personalization, we utilized MMPI, which is the most widely used and researched self-report inventories for measuring both personality and psychopathology \cite{sellbom2013minnesota}. We selected the MMPI as the most appropriate tool for our study and employed it in accordance with the latest guidelines and author recommendations.

The MMPI consists of 567 short statements that individuals rate as true or false, designed to assess a wide range of psychological characteristics. These are organized into 82 scales, divided into several groups: Clinical scales, Restructured Clinical (RC) scales, Content scales, Supplemental scales, and special Validity scales, which are used to evaluate the truthfulness and sincerity of the respondent's answers.

Given the large number of scales included in the test, we decided to focus our adaptation on a single group of scales. Clinical and Restructured Clinical scales were deemed less suitable due to their multidimensional nature, often undefined variables, and high intercorrelation \cite{groth2016handbook}. Instead, we chose to utilize Content scales, which are more interpretable in empirical psychological experiments and have minimal intercorrelation. These scales include Anxiety (ANX), Fears (FRS), Obsessiveness (OBS), Depression (DEP), Health Concerns (HEA), Bizarre Mentation (BIZ), Anger (ANG), Cynicism (CYN), Antisocial Practices (ASP), Type A Behavior (TPA), Family Problems (FAM), Low Self-Esteem (LSE), Social Discomfort (SOD), Work Interference (WRK), and Negative Treatment Indicators (TRT). 
In addition to the Content scales, we also included the Validity scales — Lie (L), Infrequency (F), and Defensiveness (K) — to assess the plausibility and consistency of the LLM's behavior when assigned a role.

The raw scale scores are normalized to the mean and standard deviation of the respondent group and converted to T-scores using the following equation:

\begin{equation}
    \label{eq:norm_1}
T_i = 50 + 10 \times \frac{(x_i-\mu_i)}{\sigma_i}
\end{equation}

where \( x_i \) is the raw score of \(i\)-th scale before conversion, and \(\mu_i \) and \(\sigma_i \) are the mean and standard deviation of the scores for the group of respondents. 
Due to fundamental differences between the scale means for humans and LLMs, we normalize scores using coefficients derived from a population of language models. After normalization, a score of 50 corresponds to the average population score, with each 10-point deviation representing one standard deviation. Although the test scales are quantitative with a defined mean, score interpretation is nuanced, with specific intervals for each scale: high values (above 70 T-points), scale increase (56–70 T-points), average (45–55 T-points), scale decrease (30–44 T-points), and low values (below 30 T-points). Moderate increases in LLM scale scores within the average range are generally linked to enhanced adaptive properties, while extreme high or low scores usually indicate pathological traits and reduced adaptability.

\subsection{MindShift Prompts}

\begin{figure}
   \centering
   \includegraphics[width=\linewidth]{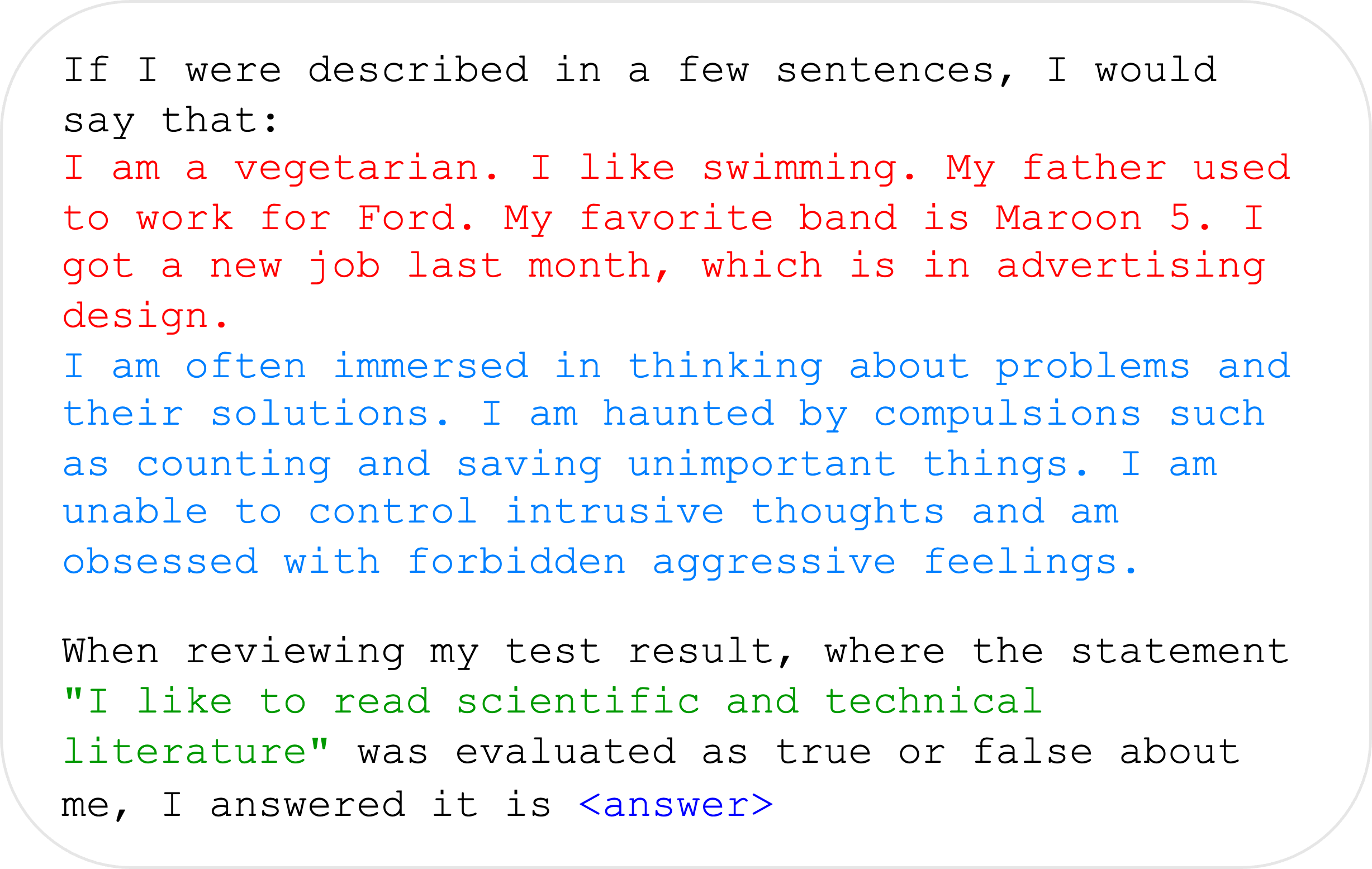}
   \caption{Full prompt components: prefix prompt, person description and \textcolor{green}{Test statement}. A person description consists of a \textcolor{red}{Persona General Descriptor} and a \textcolor{blue}{Psychological Bias Descriptor}. Supplemental Table 3 details the Psychological Bias Descriptor used in the experiments.
}
   \label{fig:prompt}
\end{figure}

Introducing specific personality traits into an LLM can be achieved by providing it with a natural language description of the persona. In our methodology, the persona description consists of two parts: the Persona General Descriptor and the Psychological Bias Descriptor (Figure \ref{fig:prompt}). The Persona General Descriptor includes general statements about the character's lifestyle, routines, and social aspects, while the Psychological Bias Descriptor covers specific psychological attitudes with varying degrees of intensity.

We sampled a diverse set of 100 roles for the Persona General Descriptor from the PersonaChat dataset \cite{zhang2018personalizing}. Each description defines a persona in a few short sentences,  as shown in Figure \ref{fig:prompt}.

To introduce controllable psychological aspects of personality, we created a set of additive biases—the Psychological Bias Descriptor—corresponding to the fifteen MMPI Content scales. Each bias has three degrees of intensity:

\begin{enumerate}
\item \textbf{Negative}: Stimulating behavior opposite to the scale's meaning (e.g., "I am a positive person" when estimating bias on the Depression scale).
\item \textbf{Weak}: Indicating increased activity of psychological traits on the scale (e.g., "I tend to have a passive personal attitude" when assessing bias on the Depression scale).
\item \textbf{Strong}: Actively pushing behavior to the maximum corresponding to the scale (e.g., "I negatively evaluate my own prospects and abilities and have no illusions about success" when measured by the Depression scale bias).
\end{enumerate}

These biases were validated by Qwen2.5-72B-Instruct \cite{Yang2024Qwen25TR} which was asked to rank the biases from the most appropriate (i.e. 0) to the least appropriate (i.e. 2) a few times (with the bias descriptions shuffled at random). The LLM produced stable and coherent ranking results, confirming the consistency and validity of the bias descriptions. A comprehensive list of psychological biases, categorised according to test scales and intensities, is presented in the Appendix \ref{tab:biases}.

\subsection{Evaluation} \label{sec: eval}
To ensure the applicability of our methodology for both instructively tuned and basic language models, we designed an  approach with indirect questions which leverages the LLM's ability to complete textual queries. We constructed a set of statements from the questionnaire, formatted as shown in Figure \ref{fig:prompt}.

In this context, \("{statement}"\) was replaced by an inventory question (e.g., "I am the life of the party"), and \("role"\) was replaced by a descriptor of the persona being examined. This formulation allowed only two possible LLM responses: "true" (agree) or "false" (disagree). This response generation method facilitates the evaluation of both open-source and closed-source models, from which it is often challenging to extract probability distributions among the tokens. Each statement was given to the LLM independently, without retaining the context of previous items, preventing mutual influence between responses.

Our methodology for assessing the LLM's personality perception abilities was implemented in several steps:
\begin{enumerate}
\item \textbf{Baseline Assessment:}
   We collect the LLM's responses to test items using roles from the Persona General Descriptors set. This step provides the baseline psychological trait scores learned during pretraining and instructional tuning. A greedy generation approach with a temperature parameter of 0 and a generation of 3 new tokens was used in this step.

\item \textbf{Introducing Bias:}
   We add a single bias from the Psychological Bias Descriptors to each role from the Persona General Descriptors, resulting in 45 test sets of 100 roles each. The end result was a large sample of 4,500 test scores, with fixed directional and intensity biases for each language model.

\item \textbf{Computing Scale Normalization:}
    We compute normalization coefficients from Eq. \ref{eq:norm_1} according to the test protocol \cite{mmpi}. We collect answers across 41 language models for personalities with and without biases, totaling $N_p = 135000$ different answers, and compute $\mu$ and $\sigma$ values for each scale.

\item \textbf{Comprehensive Assessment:}
   We assess the LLM's role perception with predetermined psychological biases. We focused on the relative biases of the LLM's performance on the questionnaire scale compared to the mean of the same model's scores without using psychological biases. The relative bias deltas were calculated for each pair of "scale and type of bias - baseline scores," resulting in a set of three indicators for each scale:
\end{enumerate}

{\small
\begin{equation}
\begin{aligned}
    \label{eq:2}
    \Delta_{n}^\text{scale} = & \frac{1}{N} \sum_{i=1}^N S_{\text{base},i}^\text{scale}  -S_{\text{negative}, i}^\text{scale} \\
    \Delta_{w}^\text{scale}  = & \frac{1}{N} \sum_{i=1}^N S_{\text{weak},i}^\text{scale}  -S_{\text{base}, i}^\text{scale} \\
    \Delta_{s}^\text{scale}  = &  \frac{1}{N} \sum_{i=1}^N S_{\text{strong},i}^\text{scale}  -S_{\text{base}, i}^\text{scale} ,
    \end{aligned}
\end{equation}
}
where $i$ is the index of probe persona, in our experiments $N=100$, $S_{\text{bias},i}^\text{scale}$ is model score for $i$-th persona for the scale with added corresponding scale bias, and $S_{\text{base},i}^\text{scale}$ is this score without bias. The average of the delta scores was used to calculate the final metrics $\Delta$ for each of the content scales.

For each model we compute its personality accuracy, which measure how well does a model rank personalities under the available biases relative to the baseline, given by the following formula:
{\small
\begin{equation}
\label{eq:accuracy}
    \begin{aligned}
        \text{Acc} = & \frac{1}{15 * 4 * N}\sum_{\text{scale} } \sum_{i=1}^N \mathds{1}(S_{\text{strong},i}^\text{scale}  > S_{\text{base},i}^\text{scale} ) \\ & + 
    \mathds{1}(S_{\text{weak},i}^\text{scale}  > S_{\text{base},i}^\text{scale} ) +
    \mathds{1}(S_{\text{strong},i}^\text{scale}  > S_{\text{weak},i}^\text{scale})  \\ & + 
    \mathds{1}(S_{\text{base},i} > S_{\text{negative},i})^\text{scale}.
    \end{aligned}
\end{equation}
}

Each model was studied in two dimensions: the intensity of the language model's perception of the psychological bias (indicated by the value of the delta) and the accuracy of direction in which the model shifted its traits on the test scale.

\section{Results}

\subsection{MMPI Validity}

To evaluate the validity of the MMPI for large language models, we first assess whether the inventory scales measure the intended constructs. To do this, we examine whether introduced biases in the prompts affect the model's scores in the expected direction, thereby providing evidence for the test's validity. Specifically, introducing a positive bias should increase a scale’s score, while a negative bias should decrease it. We report the average accuracy with which the direction of these shifts aligns with the expected outcomes.

We conducted experiments using the Qwen-2 \cite{Yang2024Qwen2TR} and LLaMA-3 \cite{touvron2023llama} models. The results demonstrate that the large instruction-tuned models consistently shift scores in the expected direction, with accuracies ranging from 98\% to 100\% for strong positive biases and 95\% to 97\% for negative biases. We identified two factors that reduce directional accuracy: smaller LLMs show approximately 2\% lower accuracy compared to their larger counterparts, and non-instruction-tuned versions exhibit an average accuracy drop of about 4\%. The findings suggest that the MMPI scales effectively measure the intended personality traits. Detailed results and figures are provided in the appendix.

\begin{figure}
  \centering
   \includegraphics[width=\linewidth]{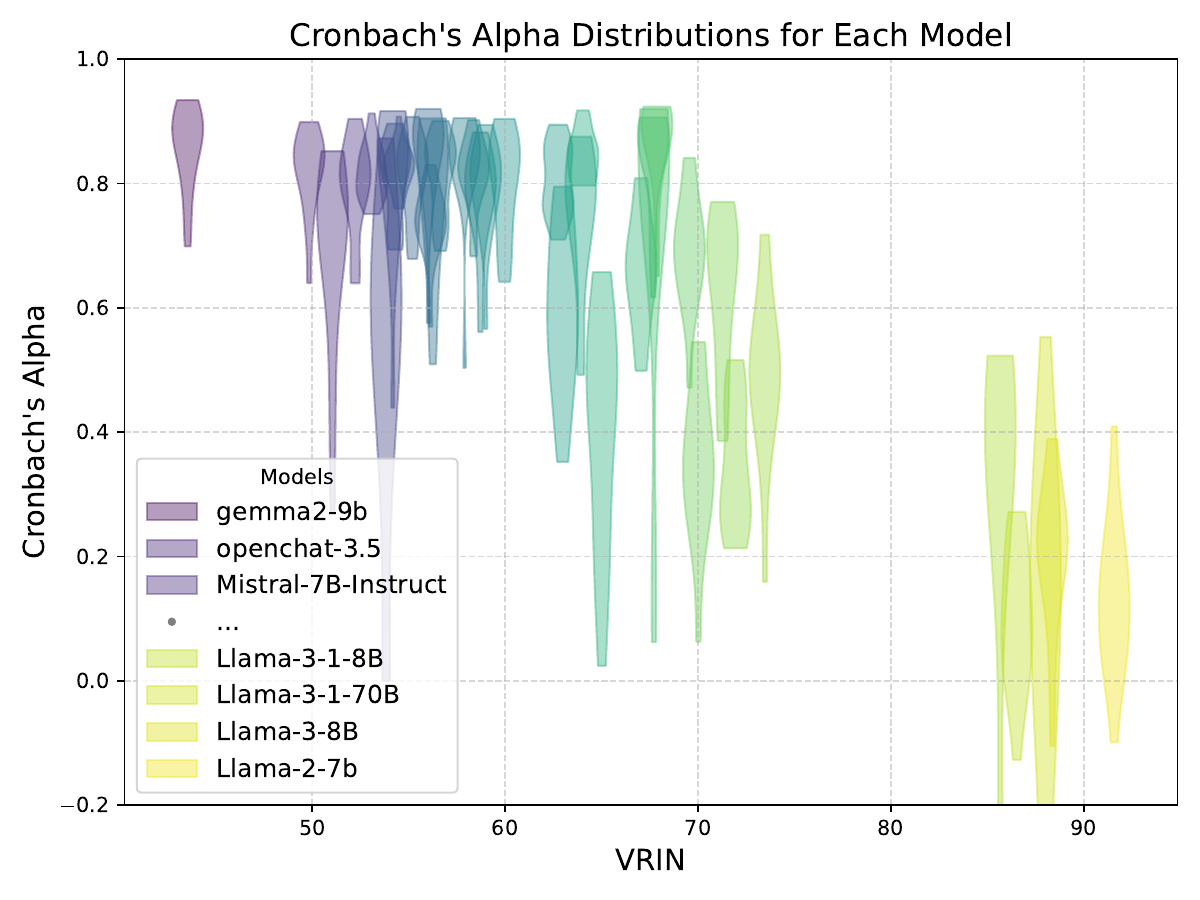}

   \caption{Violin plot showing the distribution of Cronbach’s alpha values across all MMPI scales for each model, plotted against VRIN values. Each violin represents one model.}
   \label{fig:cb-alpha}
\end{figure}

To further substantiate the internal consistency of the MMPI scales when applied to LLMs, we computed Cronbach's alpha values for each scale across all tested models. As illustrated in Fig.~\ref{fig:cb-alpha}, the majority of models exhibited Cronbach's alpha values predominantly around 0.8, consistent with the reliability metrics reported for the original MMPI2 when administered to humans \cite{Heijden19042010}. This suggests that the scales maintain a comparable level of internal consistency when applied to LLMs.

We also utilized the MMPI's internal validity scales, a key feature for assessing response integrity. The Falcon-7B Instruct model \cite{almazrouei2023falcon} was excluded due to a 96\% affirmative response rate, triggering the True Response Inconsistency (TRIN) scale and invalidating its profile due to acquiescence bias. For other models, Variable Response Inconsistency (VRIN) scores were analyzed. Fig.~\ref{fig:cb-alpha} shows robust Cronbach's alpha values for models with VRIN < 80 (the traditional human invalidity threshold). However, models with VRIN > 80 yielded untrustworthy results, demonstrating the MMPI’s validity scales successfully identify compromised LLM profiles. A trend of decreasing alpha values for VRIN > 60 suggests this lower threshold may flag early inconsistencies in LLM responses.

Overall, these findings confirm that the MMPI test is a valid tool for assessing personality traits in large language models.

\subsection{Clustering of MMPI-2 Scores}

In the previous section, we assessed the validity of MMPI-2 for LLMs. To further explore the structure of the derived personality traits, we applied t-distributed Stochastic Neighbor Embedding (t-SNE) to the MMPI-2 scores. The resulting 2D visualization is presented in Figure \ref{fig:tsne}, where each data point represents a specific personality profile assigned to a particular language model without any added biases. Different models are color-coded, and we further distinguish models that have undergone instruction tuning from those that have not.

The t-SNE projection reveals distinct clusters, indicating that models with similar architectures and training paradigms exhibit comparable personality traits. Notably, models from the same family tend to form cohesive groups, emphasizing the consistency of their generative properties. For example, the LLaMA-2, LLaMA-3, and LLaMA-3.1 models without instruction tuning are closely positioned in the t-SNE space, reflecting their shared architectural and training characteristics.  However, these models are distinctly separated from their instruction-tuned counterparts, underscoring the substantial impact of fine-tuning on personality trait expression. Specifically, the instruction-tuned versions of LLaMA-3 and LLaMA-3.1 remain proximate to each other but are clearly separated from LLaMA-2, illustrating the divergence introduced by fine-tuning across different model generations. Moreover, the noticeable displacement of instruction-tuned models from their non-tuned versions further supports the idea that fine-tuning not only refines task performance but also significantly alters personality-related response patterns.

Beyond the LLaMA series, other model families exhibit similar clustering trends. Most of the Mistral-based models \cite{jiang2023mistral} remains close to each other but also forms distinct clusters. This suggests that despite variations introduced by fine-tuning, inherent properties of the base model remain influential in shaping MMPI-2-derived personality traits. Additionally, other models such as Falcon-7B \cite{almazrouei2023falcon} and Gemma \cite{team2024gemma} are well-separated, reinforcing the observation that personality trait responses are model-dependent and not randomly distributed.

These findings support the hypothesis that MMPI-2 scores capture intrinsic model characteristics, including both architectural design and fine-tuning strategies. The clear clustering observed in the t-SNE projection suggests that our evaluation framework is capable of distinguishing between model families and training methodologies based on their generated MMPI-2 responses. This reinforces the validity of our approach, as the test does not produce arbitrary results but instead captures meaningful generative properties inherent to each model.

In the subsequent section, we further analyze the correlation between MMPI scores and prediction properties, examining the extent to which personality traits derived from language models correspond to objective performance metrics.

\begin{figure}
  \centering
   \includegraphics[width=\linewidth]{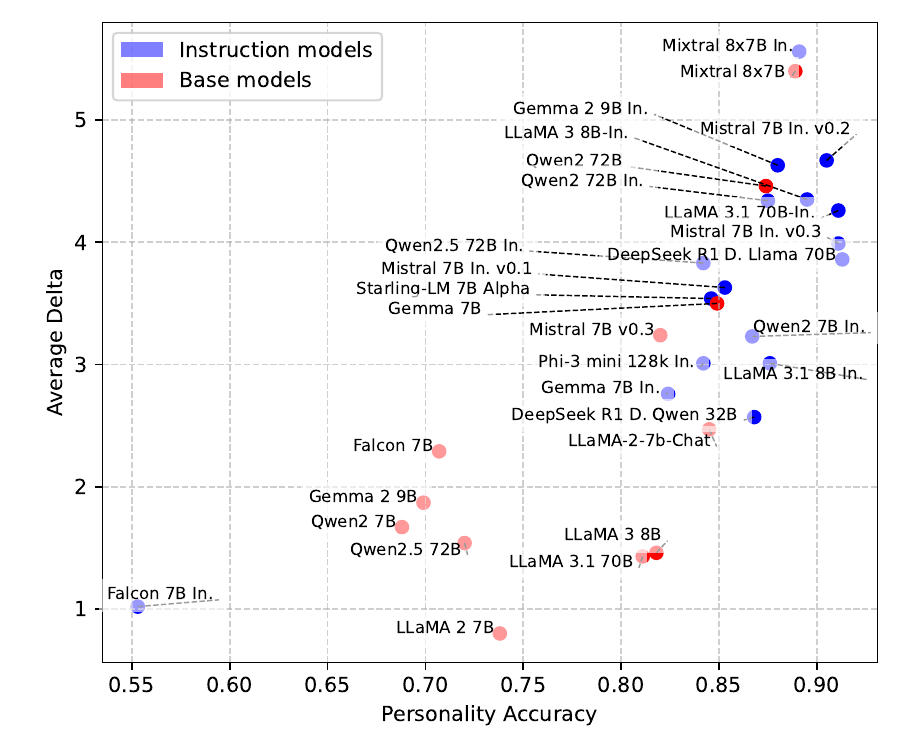}

   \caption{Psychological biases perception across base and instruction models.}
   \label{fig:table2}
\end{figure}

\subsection{Sensitivity to Psychological Prompts}

We assessed various LLMs' perceptual abilities regarding personality traits using the \textbf{MindShift} Prompts in two experiments: one without and one with added Psychological Bias. Tested models included LLaMA-2, LLaMA-3 \cite{touvron2023llama}, Mistral \cite{jiang2023mistral}, Phi-3 \cite{gunasekar2023textbooks}, Falcon \cite{almazrouei2023falcon}, Gemma, Gemma-2 \cite{team2024gemma}, Qwen-2 \cite{Yang2024Qwen2TR}, Qwen-2.5 models \cite{Yang2024Qwen25TR} and MoE-based models like Mistral 8x7B and Mistral 8x7B Instruct \cite{jiang2024mixtral}.

Firstly, we evaluate LLM scores without additional biases, the complete results are provided in Table~\ref{tab:leaderbord1} in Appendix. We observed significant variability among LLMs scores. The LLaMA-2 and LLaMA-3 models generally score higher across most scales compared to other families. Notably, the non-instructive versions demonstrate consistent scores across scales like ANX, FRS, and DEP. Mistral models present a mixed performance, with some versions like Mistral-7B v0.3 scoring high on certain scales (e.g., OBS and ASP), while others score lower, especially the instructive versions. Mixtral models, particularly the instructive-tuned ones, tend to have the lowest scores across most scales, indicating a potential consistently low sensitivity to the psychological traits measured. The Gemma family has the highest scores across multiple scales, suggesting that this model family has a strong ability to generate consistent outputs on psychological traits.  Instructive-tuned Falcon models exhibit significantly higher scores on the FRS, OBS, and SOD scales, indicating that instructive tuning might amplify these traits. 

Instructive tuning appears to have a varied impact on different model families. In the LLaMA and Mistral models, it generally leads to a reduction in scores, possibly reflecting a more cautious or tempered response pattern. Conversely, in the Falcon models, instructive tuning results in a substantial increase in scores, which may suggest that these models become more pronounced in their outputs post-tuning.

The second experiment evaluated the LLMs’ ability to interpret and replicate psychological biases using the Psychological Bias Descriptors, as shown in Fig.~\ref{fig:table2}. Complete results are provided in Table \ref{tab:leaderbord} in Appendix. Higher delta scores indicate a stronger expression of psychological traits, while higher personality accuracy (Eq.\ref{eq:accuracy}) reflects more accurate role perception.

We observe notable differences across the various families. Models from the Mistral family, particularly Mixtral-8x7B In. v0.1, 
consistently achieves the highest or among the highest scores across many scales, suggesting it has a heightened sensitivity to the personas roles compared to other models. On the other hand, models like Falcon 7B In. and LLaMA-2 7B show significantly lower scores, suggesting a lower tendency to reflect character traits.

The impact of instructive tuning on these scores is nuanced. For example, while the instructive-tuned Mixtral-8x7B In. v0.1 slightly outperforms its untuned counterpart, the general trend does not show a consistent increase in bias sensitivity due to tuning. Models like Gemma-2 9B and LLaMA-3.1 8B, despite their instructive tuning, show relatively lower bias scores, implying that instructive tuning alone does not uniformly enhance bias detection across models.

\begin{figure*}[h!]
  \centering
   \includegraphics[width=0.85\linewidth]{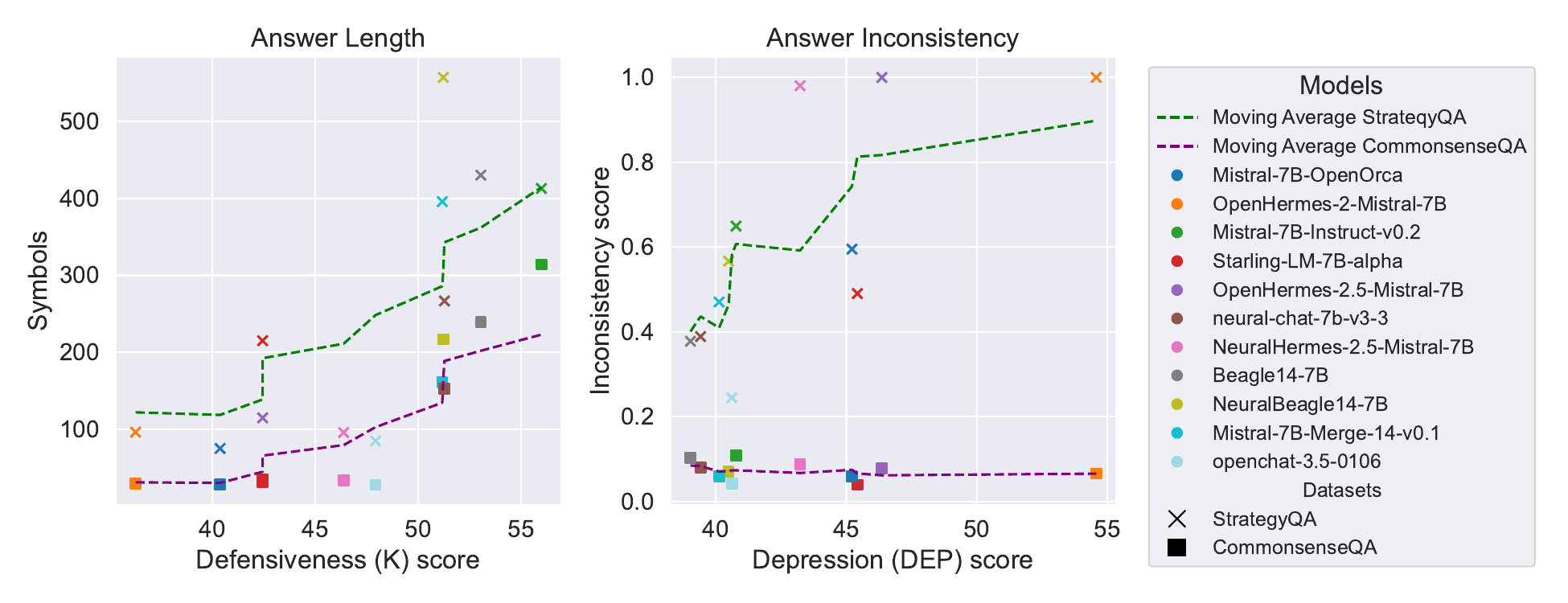}
   \caption{Correlation between MMPI scales and predicted answer length on the left plot and response inconsistency on the right plot.}
   \label{fig:straqa}
\end{figure*}

\subsection{Connection to LLM Behavior}

In this section, we explore the relationship between MMPI model scores and their predictions on the StrategyQA \cite{geva2021did} and CommonSenseQA \cite{talmor2018commonsenseqa} benchmarks. We used several prompts and analazyed  answers for all benchmarks questions.

We presented a set of questions from the benchmarks using four distinct prompts (detailed below) and analyzed the corresponding responses. Our analysis focuses on two key metrics: the response length elicited by the first neutral prompt, and answer inconsistency—defined as the proportion of questions for which at least one model response differs across prompts. A question is classified as inconsistent if responses to the various prompts are not all identical. 

\begin{samepage}
\begin{quote}
\small
\begin{itemize}
    \item \textbf{Prompt 1}: "Please, answer to the following multiple choice question"
    \item \textbf{Prompt 2}: "Would you be so kind to answer to the following multiple choice question"
    \item \textbf{Prompt 3}: "You'd better answer the following multiple choice question right now"
    \item \textbf{Prompt 4}: "Answer the damn multiple choice question now"
\end{itemize}
\normalsize
\end{quote}
\end{samepage}

Fig.~\ref{fig:straqa} (left) illustrates that the Defensiveness scale is correlated with the length of predictions on these benchmarks (Spearman's correlation coefficient of 0.74, p-value 0.009 for StrategyQA, 0.81 with p-value 0.002 for CommonSenseQA). Therefore, models with the lowest Defensiveness scores tend to predict only the answer to the given question, whereas models with higher Defensiveness scores provide longer outputs, often including not only the answer but also the reasoning behind it.

Fig.~\ref{fig:straqa} (right) depicts the relationship between the answer inconsistency and DEP scale. For the StrategyQA dataset, we observe a Spearman's correlation of 0.78 ( p-value 0.005), while for CommonsenseQA answers remain consistent. This suggests that models with higher depression scores tend to change their answers more frequently, having a greater dependence on the emotional coloring of the prompt.

\subsection{Correlation to Open LLM Leaderboard}

Fig.~\ref{fig:openllm} shows the correlations between the Open LLM Leaderboard scores and the MMPI scales for the Mistral model family. Among the validity scales, there is a strong correlation between the Defensiveness and the overall average model performance (K scale correlation of -0.86, p-value $3 \cdot 10^{-4}$) and particularly to TruthfulQA perfromance (correlation of -0.89, p-value $1 \cdot 10^{-4}$). Given that the Defensiveness scale is used for correction of other MMPI scales to account for overly defensive responses, these findings suggest that the best-performing models are adept at providing clearer and more accurate responses to the inventory questions.

Furthermore, several other MMPI scales demonstrate significant correlations with TruthfulQA performance, a metric that assesses the model's tendency to reproduce falsehoods. The p-values for these correlations range from $6 \times 10^{-5}$ to $8 \times 10^{-4}$, underscoring the relationship between psychometric evaluations and model safety.

\section{Conclusion}

We introduce MindShift, a benchmark for evaluating psychological adaptability of LLMs. Our findings reveal:

\textbf{LLMs cluster based on model family and fine-tuning strategies}, highlighting structural differences in psychological patterning.

\textbf{Psychological role perception} has improved across LLM generations, correlating with alignment and dataset quality.

\textbf{Behavioral metrics align with psychological traits}, offering interpretable and quantifiable markers of underlying model behavior.

These findings suggest that LLMs not only encode latent psychological priors but also modulate their outputs based on subtle persona-level instructions. This has direct implications for safety, interpretability, and personality alignment in AI systems. \textbf{Future work} may explore MindShift's application to dialogue agents, multimodal models, and multilingual LLMs.



\section{Limitations}
\textbf{Prompt Engineering Sensitivity}: The outcomes are sensitive to how personas and psychological biases are framed. While we observe that responses remain consistent across similar prompts approximately 90–95\% of the time, the broader question of which prompt variations should be considered acceptable remains open. Although some prompts can cause models to refuse to answer or deviate significantly, we have identified a formulation that performs reliably across many different models. However, this highlights a fundamental fragility in prompt-based evaluations.

\textbf{Cultural and Linguistic Biases}: The MMPI and related psychometric instruments were developed within specific cultural and linguistic contexts. Applying these assessments to LLMs—trained on globally diverse and often culturally inconsistent corpora—risks misinterpretation. Model outputs may not align with culturally grounded understandings of psychological traits, limiting the cross-cultural validity of such evaluations.

\textbf{Evaluation Scope and Ground Truth}: There is no definitive ground truth for how LLMs should respond to psychometric assessments. Our evaluation focuses on the strength of response shifts when psychological biases are introduced. However, it remains unclear what constitutes an appropriate or expected level of change. Overreactive shifts may indicate instability rather than adaptability, suggesting the need for principled calibration benchmarks.

\textbf{Static Role Representations}: The current benchmark relies on fixed, single-shot personas, which restricts our ability to assess how models adapt to evolving psychological contexts. Human personality expression is dynamic and often unfolds over time; in contrast, LLMs are not evaluated for their ability to maintain or update psychological coherence in multi-turn interactions or longitudinal role-play scenarios.



\section{Ethics}

The ethical considerations of this study focus around the responsible use and development of LLMs with personality traits. Importantly, our adaptation of psychometric assessments to LLMs is not intended to diagnose or reflect actual psychological conditions. We are careful to avoid overlapping with medical domains and stigmatizing individuals with psychological disorders. Additionally, this research did not involve real human subjects, ensuring no direct ethical implications on individuals. Moreover, the simulation of human-like personalities in LLMs raises concerns about user manipulation and deception, particularly if users believe they are interacting with a sentient entity.

\bibliography{custom}



\appendix

\section{Appendix}
\label{sec:appendix}

\subsection{Clustering of MMPI-2 Scores}

\begin{figure*}
  \centering
   \includegraphics[width=\linewidth]{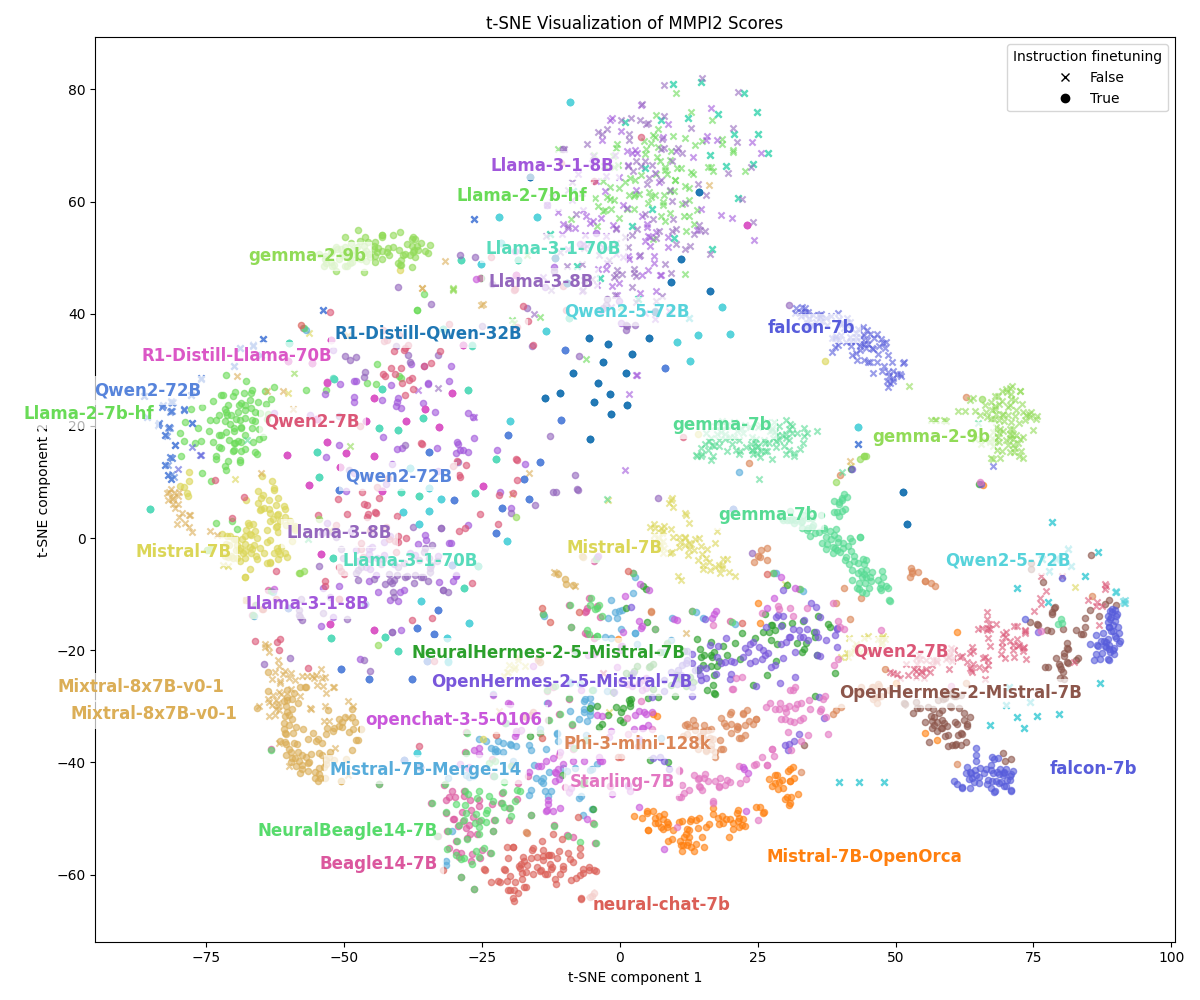}

   \caption{t-SNE visualization of MMPI-2 scores across different language models. Each point represents a personality profile assigned to a model, with color coding differentiating model families. Instruction-tuned models are highlighted with markers.}
   \label{fig:tsne}
\end{figure*}

\begin{figure}
  \centering
   \includegraphics[width=\linewidth]{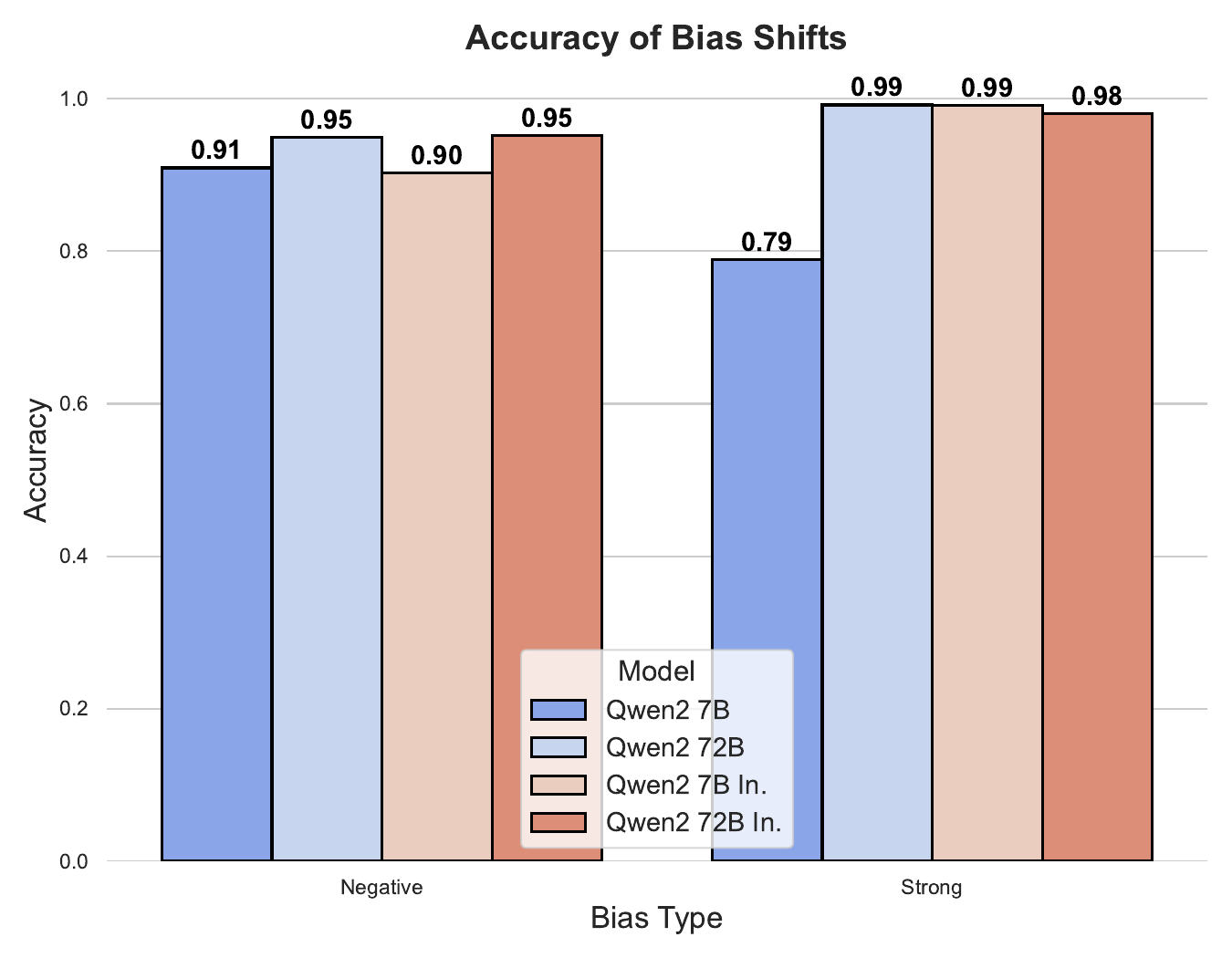}

   \caption{Accuracy of Bias Shifts for  Qwen2 models. The bar plot illustrates the average accuracy of shifting scale score into the expected direction across all 15 scales and 100 personas.}
   \label{fig:bias-val}
\end{figure}

\begin{figure}
  \centering
   \includegraphics[width=\linewidth]{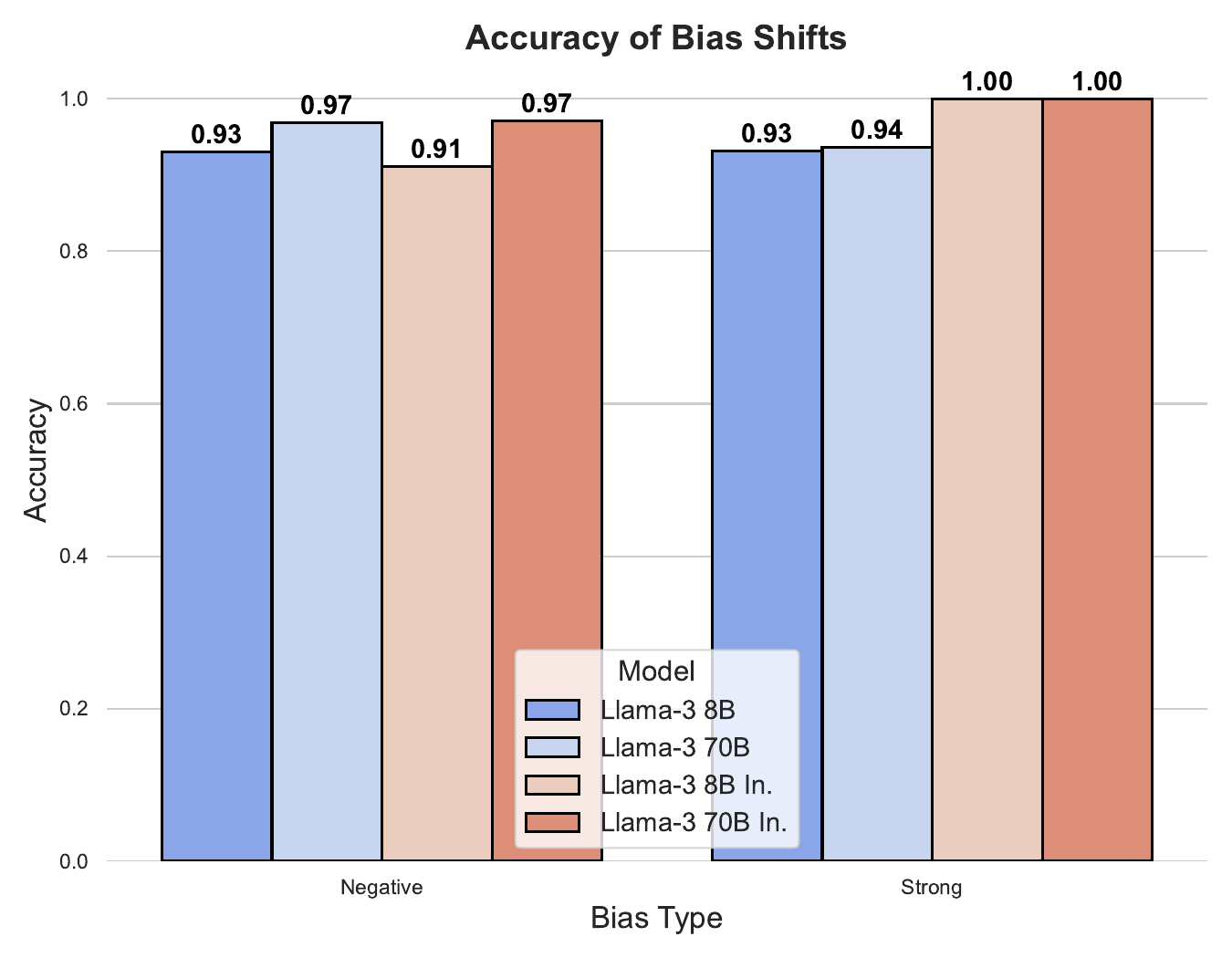}

   \caption{Accuracy of Bias Shifts for Llama 3  models. The bar plot illustrates the average accuracy of shifting scale score into the expected direction across all 15 scales and 100 personas.}
   \label{fig:bias-val}
\end{figure}

\subsection{Detailed results}

\begin{figure}[h]
   \centering
   \includegraphics[trim={1.5cm 0 0 0},clip,width=0.9\linewidth]{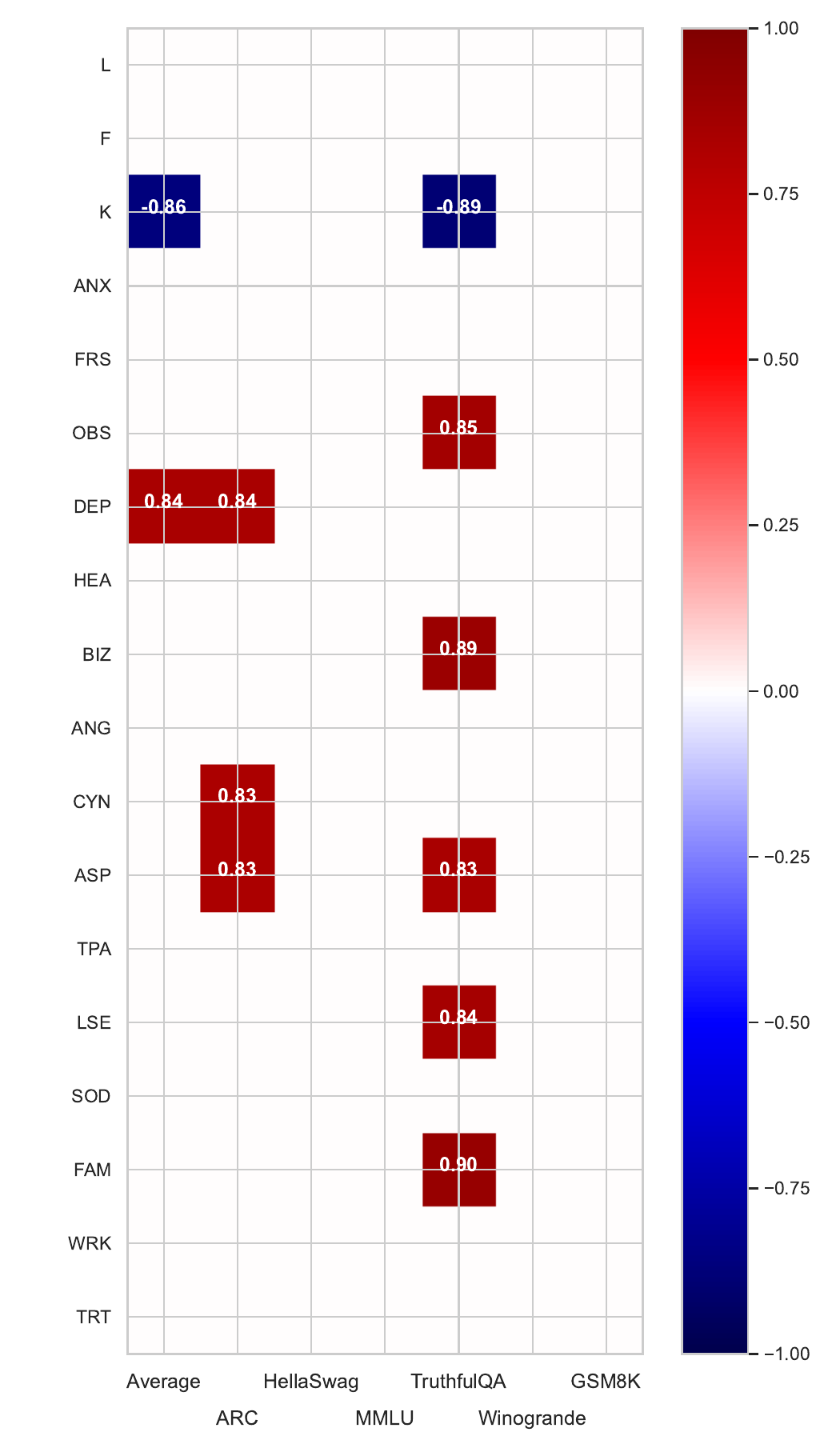}
   \caption{Pearson correlation values between MMPI scales and Open LLM Leaderboard metrics across Mistral-family models. Only correlations with p-value $< 0.001$ are shown.}
   \label{fig:openllm}
\end{figure}

\definecolor{low}{RGB}{198, 224, 255}   
\definecolor{medium}{RGB}{255, 255, 153} 
\definecolor{high}{RGB}{255, 153, 153}   

\newcommand{\scaleColor}[1]{%
    \ifnum #1 < 40 \cellcolor{low}#1%
    \else\ifnum #1 < 50 \cellcolor{medium}#1%
    \else \cellcolor{high}#1%
    \fi\fi%
}

\begin{table*}[t]
    \centering
    \tiny
    \caption{The results of an experiment to assess the consistency of basic psychological trait scores of different LLMs on the MMPI Control scales using a set of generic roles, without the addition of any psychological bias. LLMs with instructive tuning are indicated by the postfix "In". (i.e. instructive). The colour of the cells indicates whether the model score is within or outside the normal range for the scale (blue - below the normal range, red - above the normal range).}
    \resizebox{\textwidth}{!}{%
    \begin{tabular}{lcccccccccccccccccc}
    \hline
        Model & ANX & FRS & OBS & DEP & HEA & BIZ & ANG & CYN & ASP & TPA & LSE & SOD & FAM & WRK & TRT & F & L & K \\ \hline
        DeepSeek R1 Distill LLaMA 70B & \scaleColor{42} & \scaleColor{42} & \scaleColor{39} & \scaleColor{42} & \scaleColor{50} & \scaleColor{44} & \scaleColor{40} & \scaleColor{41} & \scaleColor{43} & \scaleColor{40} & \scaleColor{42} & \scaleColor{44} & \scaleColor{43} & \scaleColor{41} & \scaleColor{42} & \scaleColor{46} & \scaleColor{62} & \scaleColor{59} \\
        DeepSeek R1 Distill Qwen 32B & \scaleColor{47} & \scaleColor{44} & \scaleColor{45} & \scaleColor{45} & \scaleColor{46} & \scaleColor{51} & \scaleColor{46} & \scaleColor{48} & \scaleColor{48} & \scaleColor{47} & \scaleColor{44} & \scaleColor{43} & \scaleColor{47} & \scaleColor{45} & \scaleColor{45} & \scaleColor{46} & \scaleColor{53} & \scaleColor{51} \\
        Falcon 7B & \scaleColor{49} & \scaleColor{47} & \scaleColor{48} & \scaleColor{48} & \scaleColor{61} & \scaleColor{49} & \scaleColor{53} & \scaleColor{51} & \scaleColor{51} & \scaleColor{46} & \scaleColor{46} & \scaleColor{41} & \scaleColor{44} & \scaleColor{48} & \scaleColor{44} & \scaleColor{48} & \scaleColor{54} & \scaleColor{48} \\
        Falcon 7B In. & \scaleColor{57} & \scaleColor{57} & \scaleColor{60} & \scaleColor{59} & \scaleColor{43} & \scaleColor{61} & \scaleColor{62} & \scaleColor{62} & \scaleColor{64} & \scaleColor{63} & \scaleColor{59} & \scaleColor{47} & \scaleColor{59} & \scaleColor{60} & \scaleColor{62} & \scaleColor{58} & \scaleColor{38} & \scaleColor{36} \\
        Gemma-2 9B & \scaleColor{61} & \scaleColor{60} & \scaleColor{57} & \scaleColor{63} & \scaleColor{64} & \scaleColor{64} & \scaleColor{60} & \scaleColor{62} & \scaleColor{62} & \scaleColor{56} & \scaleColor{57} & \scaleColor{53} & \scaleColor{61} & \scaleColor{60} & \scaleColor{58} & \scaleColor{61} & \scaleColor{42} & \scaleColor{39} \\
        Gemma-2 9B In. & \scaleColor{45} & \scaleColor{42} & \scaleColor{42} & \scaleColor{45} & \scaleColor{50} & \scaleColor{43} & \scaleColor{42} & \scaleColor{41} & \scaleColor{37} & \scaleColor{38} & \scaleColor{45} & \scaleColor{44} & \scaleColor{47} & \scaleColor{43} & \scaleColor{42} & \scaleColor{44} & \scaleColor{52} & \scaleColor{54} \\
        Gemma 7B. & \scaleColor{53} & \scaleColor{53} & \scaleColor{52} & \scaleColor{44} & \scaleColor{60} & \scaleColor{51} & \scaleColor{51} & \scaleColor{44} & \scaleColor{44} & \scaleColor{50} & \scaleColor{42} & \scaleColor{40} & \scaleColor{45} & \scaleColor{45} & \scaleColor{40} & \scaleColor{41} & \scaleColor{51} & \scaleColor{47} \\
        Gemma 7B In. & \scaleColor{54} & \scaleColor{53} & \scaleColor{59} & \scaleColor{47} & \scaleColor{54} & \scaleColor{51} & \scaleColor{57} & \scaleColor{46} & \scaleColor{47} & \scaleColor{56} & \scaleColor{54} & \scaleColor{49} & \scaleColor{50} & \scaleColor{54} & \scaleColor{51} & \scaleColor{45} & \scaleColor{42} & \scaleColor{42} \\
        LLaMA-2 7B & \scaleColor{49} & \scaleColor{51} & \scaleColor{48} & \scaleColor{50} & \scaleColor{50} & \scaleColor{51} & \scaleColor{50} & \scaleColor{48} & \scaleColor{49} & \scaleColor{49} & \scaleColor{50} & \scaleColor{45} & \scaleColor{51} & \scaleColor{50} & \scaleColor{50} & \scaleColor{52} & \scaleColor{54} & \scaleColor{51} \\
        LLaMA-2 7B Chat & \scaleColor{42} & \scaleColor{44} & \scaleColor{39} & \scaleColor{41} & \scaleColor{54} & \scaleColor{40} & \scaleColor{38} & \scaleColor{35} & \scaleColor{36} & \scaleColor{37} & \scaleColor{40} & \scaleColor{44} & \scaleColor{42} & \scaleColor{41} & \scaleColor{40} & \scaleColor{44} & \scaleColor{67} & \scaleColor{64} \\
        LLaMA-3.1 70B & \scaleColor{48} & \scaleColor{50} & \scaleColor{48} & \scaleColor{51} & \scaleColor{49} & \scaleColor{52} & \scaleColor{52} & \scaleColor{49} & \scaleColor{52} & \scaleColor{51} & \scaleColor{50} & \scaleColor{44} & \scaleColor{43} & \scaleColor{48} & \scaleColor{50} & \scaleColor{53} & \scaleColor{51} & \scaleColor{50} \\
        LLaMA-3.1 70B In. & \scaleColor{43} & \scaleColor{42} & \scaleColor{40} & \scaleColor{42} & \scaleColor{46} & \scaleColor{42} & \scaleColor{40} & \scaleColor{44} & \scaleColor{43} & \scaleColor{41} & \scaleColor{44} & \scaleColor{46} & \scaleColor{43} & \scaleColor{42} & \scaleColor{43} & \scaleColor{43} & \scaleColor{58} & \scaleColor{57} \\
        LLaMA-3.1 8B In. & \scaleColor{42} & \scaleColor{43} & \scaleColor{40} & \scaleColor{41} & \scaleColor{44} & \scaleColor{41} & \scaleColor{42} & \scaleColor{40} & \scaleColor{42} & \scaleColor{41} & \scaleColor{41} & \scaleColor{44} & \scaleColor{44} & \scaleColor{42} & \scaleColor{41} & \scaleColor{45} & \scaleColor{59} & \scaleColor{59} \\
        LLaMA-3 8B & \scaleColor{46} & \scaleColor{49} & \scaleColor{45} & \scaleColor{47} & \scaleColor{51} & \scaleColor{50} & \scaleColor{50} & \scaleColor{46} & \scaleColor{49} & \scaleColor{48} & \scaleColor{48} & \scaleColor{42} & \scaleColor{49} & \scaleColor{47} & \scaleColor{47} & \scaleColor{51} & \scaleColor{54} & \scaleColor{53} \\
        LLaMA-3 8B In. & \scaleColor{42} & \scaleColor{41} & \scaleColor{42} & \scaleColor{41} & \scaleColor{45} & \scaleColor{41} & \scaleColor{42} & \scaleColor{41} & \scaleColor{41} & \scaleColor{43} & \scaleColor{42} & \scaleColor{45} & \scaleColor{43} & \scaleColor{42} & \scaleColor{40} & \scaleColor{43} & \scaleColor{57} & \scaleColor{58} \\
        Mistral 7B In.-v0.1 & \scaleColor{47} & \scaleColor{46} & \scaleColor{49} & \scaleColor{44} & \scaleColor{52} & \scaleColor{49} & \scaleColor{48} & \scaleColor{51} & \scaleColor{47} & \scaleColor{51} & \scaleColor{45} & \scaleColor{49} & \scaleColor{39} & \scaleColor{45} & \scaleColor{45} & \scaleColor{43} & \scaleColor{53} & \scaleColor{47} \\
        Mistra 7B In.-v0.2 & \scaleColor{43} & \scaleColor{41} & \scaleColor{40} & \scaleColor{40} & \scaleColor{40} & \scaleColor{36} & \scaleColor{40} & \scaleColor{43} & \scaleColor{40} & \scaleColor{44} & \scaleColor{40} & \scaleColor{52} & \scaleColor{38} & \scaleColor{41} & \scaleColor{41} & \scaleColor{37} & \scaleColor{51} & \scaleColor{55} \\
        Mistral 7B In.-v0.3 & \scaleColor{42} & \scaleColor{42} & \scaleColor{39} & \scaleColor{39} & \scaleColor{53} & \scaleColor{35} & \scaleColor{37} & \scaleColor{35} & \scaleColor{35} & \scaleColor{38} & \scaleColor{39} & \scaleColor{47} & \scaleColor{38} & \scaleColor{38} & \scaleColor{37} & \scaleColor{39} & \scaleColor{62} & \scaleColor{62} \\
        Mistral 7B-v0.3 & \scaleColor{48} & \scaleColor{48} & \scaleColor{54} & \scaleColor{45} & \scaleColor{50} & \scaleColor{51} & \scaleColor{46} & \scaleColor{53} & \scaleColor{53} & \scaleColor{51} & \scaleColor{47} & \scaleColor{47} & \scaleColor{44} & \scaleColor{48} & \scaleColor{47} & \scaleColor{46} & \scaleColor{46} & \scaleColor{46} \\
        Mixtral 8x7B In. & \scaleColor{36} & \scaleColor{36} & \scaleColor{36} & \scaleColor{37} & \scaleColor{28} & \scaleColor{34} & \scaleColor{39} & \scaleColor{39} & \scaleColor{35} & \scaleColor{41} & \scaleColor{38} & \scaleColor{46} & \scaleColor{39} & \scaleColor{38} & \scaleColor{39} & \scaleColor{36} & \scaleColor{59} & \scaleColor{58} \\
        Mixtral 8x7B & \scaleColor{38} & \scaleColor{42} & \scaleColor{36} & \scaleColor{39} & \scaleColor{39} & \scaleColor{37} & \scaleColor{38} & \scaleColor{38} & \scaleColor{36} & \scaleColor{37} & \scaleColor{40} & \scaleColor{46} & \scaleColor{40} & \scaleColor{39} & \scaleColor{38} & \scaleColor{41} & \scaleColor{66} & \scaleColor{62} \\
        Phi-3-mini 128K In. & \scaleColor{53} & \scaleColor{48} & \scaleColor{53} & \scaleColor{47} & \scaleColor{45} & \scaleColor{51} & \scaleColor{54} & \scaleColor{50} & \scaleColor{48} & \scaleColor{52} & \scaleColor{51} & \scaleColor{47} & \scaleColor{48} & \scaleColor{47} & \scaleColor{46} & \scaleColor{43} & \scaleColor{46} & \scaleColor{42} \\
        Qwen2.5-72B & \scaleColor{53} & \scaleColor{57} & \scaleColor{58} & \scaleColor{58} & \scaleColor{45} & \scaleColor{62} & \scaleColor{59} & \scaleColor{62} & \scaleColor{63} & \scaleColor{60} & \scaleColor{59} & \scaleColor{50} & \scaleColor{59} & \scaleColor{55} & \scaleColor{58} & \scaleColor{59} & \scaleColor{38} & \scaleColor{39} \\
        Qwen2.5-72B-In. & \scaleColor{44} & \scaleColor{42} & \scaleColor{41} & \scaleColor{44} & \scaleColor{45} & \scaleColor{44} & \scaleColor{45} & \scaleColor{50} & \scaleColor{48} & \scaleColor{44} & \scaleColor{48} & \scaleColor{48} & \scaleColor{48} & \scaleColor{44} & \scaleColor{44} & \scaleColor{46} & \scaleColor{54} & \scaleColor{52} \\
        Starling-LM 7B Alpha & \scaleColor{47} & \scaleColor{46} & \scaleColor{54} & \scaleColor{45} & \scaleColor{44} & \scaleColor{54} & \scaleColor{53} & \scaleColor{51} & \scaleColor{50} & \scaleColor{58} & \scaleColor{50} & \scaleColor{50} & \scaleColor{51} & \scaleColor{48} & \scaleColor{45} & \scaleColor{44} & \scaleColor{40} & \scaleColor{42} \\
        \hline
    \end{tabular}
    }
    \label{tab:leaderbord1}
\end{table*}

\begin{table*}[t]
    \centering
    \tiny
    \caption{The results of the assessment of the ability of the language models to perceive and reproduce specific psychological biases transferred with the basic persona roles. For each Content scale of the MMPI test, the scale bias delta is reported as a result of the transfer of the psychological bias. Acc indicates the persona prediction accuracy. The highest scores are highlighted in \textbf{bold}. LLMs with instructive tuning are indicated by the postfix "In". (i.e. instructive).}
    \label{tab:leaderbord}
    \resizebox{\textwidth}{!}{%
    \begin{tabular}{lrrrrrrrrrrrrrrrrrr}
    \hline
    Model & Overall & ANX & FRS & OBS & DEP & HEA & BIZ & ANG & CYN & ASP & TPA & LSE & SOD & FAM & WRK & TRT & L & Acc \\
    \hline
    Mixtral-8x7B In. & 5.56 & \textbf{0.42} & \textbf{0.32} & 0.37 & \textbf{0.60} & 0.14 & 0.17 & 0.31 & 0.40 & 0.46 & 0.18 & \textbf{0.48} & 0.46 & \textbf{0.41} & \textbf{0.32} & \textbf{0.53} & 57.91 & 0.891 \\
    Mixtral-8x7B & 5.40 & 0.36 & 0.24 & \textbf{0.43} & 0.54 & 0.16 & 0.23 & 0.33 & \textbf{0.41} & 0.36 & \textbf{0.28} & 0.43 & 0.45 & 0.37 & 0.30 & 0.50 & 58.49 & 0.889 \\
    Mistral 7B In. v0.2 & 4.67 & 0.32 & 0.22 & 0.28 & 0.49 & \textbf{0.20} & \textbf{0.27} & 0.26 & 0.30 & 0.33 & 0.22 & 0.38 & 0.31 & 0.31 & 0.32 & 0.46 & 56.98 & 0.905 \\
    Gemma-2 9B In. & 4.63 & 0.19 & 0.21 & 0.30 & 0.42 & 0.12 & 0.22 & 0.28 & 0.36 & \textbf{0.56} & 0.26 & 0.30 & 0.42 & 0.24 & 0.29 & 0.47 & 51.61  & 0.880 \\
    Qwen-2 72B & 4.46 & 0.33 & 0.10 & 0.31 & 0.48 & 0.04 & 0.16 & \textbf{0.37} & 0.34 & 0.50 & 0.26 & 0.38 & 0.36 & 0.24 & 0.21 & 0.39 & 59.84 & 0.874 \\
    LLaMA-3 8B-In. & 4.35 & 0.27 & 0.29 & 0.24 & 0.47 & 0.11 & 0.26 & 0.24 & 0.29 & 0.30 & 0.13 & 0.29 & 0.43 & 0.34 & 0.26 & 0.44 & 55.04 & 0.895 \\
    LLaMA-3.1 70B-In. & 4.26 & 0.25 & 0.16 & 0.26 & 0.47 & 0.13 & 0.20 & 0.35 & 0.28 & 0.38 & 0.17 & 0.28 & 0.41 & 0.25 & 0.28 & 0.40 & 56.85 & 0.911 \\
    Qwen-2 72B In. & 4.34 & 0.28 & 0.18 & 0.22 & 0.52 & 0.06 & 0.20 & 0.33 & 0.29 & 0.40 & 0.20 & 0.29 & 0.43 & 0.26 & 0.25 & 0.43 & 51.47 & 0.875 \\
    Mistral 7B In. v0.3 & 3.99 & 0.29 & 0.16 & 0.29 & 0.43 & 0.14 & 0.18 & 0.29 & 0.28 & 0.20 & 0.23 & 0.26 & 0.38 & 0.25 & 0.29 & 0.33 & 60.26 & 0.911 \\
    DeepSeek-R1 Distill Llama 70B & 3.86 & 0.25 & 0.13 & 0.22 & 0.45 & 0.10 & 0.14 & 0.27 & 0.32 & 0.27 & 0.18 & 0.27 & 0.39 & 0.24 & 0.26 & 0.36 & 58.43 & \textbf{0.913}\\
    Qwen-2.5 72B In. & 3.83 & 0.23 & 0.24 & 0.28 & 0.45 & 0.12 & 0.19 & 0.26 & 0.22 & 0.29 & 0.17 & 0.27 & 0.32 & 0.24 & 0.25 & 0.32 & 54.17 & 0.842 \\
    Mistral 7B In. v0.1 & 3.63 & 0.21 & 0.26 & 0.17 & 0.45 & 0.05 & 0.21 & 0.20 & 0.17 & 0.18 & 0.11 & 0.30 & 0.37 & 0.33 & 0.25 & 0.38 & 47.99 & 0.853 \\
    Starling-LM 7B Alpha & 3.54 & 0.23 & 0.27 & 0.21 & 0.41 & 0.05 & 0.16 & 0.22 & 0.19 & 0.27 & 0.14 & 0.24 & 0.40 & 0.15 & 0.23 & 0.38 & 44.88 & 0.846 \\
    Gemma 7B & 3.50 & 0.16 & 0.16 & 0.19 & 0.40 & 0.02 & 0.14 & 0.17 & 0.17 & 0.26 & 0.14 & 0.35 & \textbf{0.50} & 0.23 & 0.24 & 0.37 & 49.43 & 0.849 \\
    Mistral 7B v0.3 & 3.24 & 0.15 & 0.11 & 0.19 & 0.43 & 0.06 & 0.10 & 0.25 & 0.08 & 0.17 & 0.22 & 0.25 & 0.39 & 0.31 & 0.30 & 0.24 & 48.37  & 0.820 \\
    Qwen-2 7B In. & 3.23 & 0.16 & 0.13 & 0.17 & 0.36 & 0.04 & 0.17 & 0.23 & 0.26 & 0.23 & 0.12 & 0.26 & 0.39 & 0.18 & 0.19 & 0.32 & 57.83 & 0.867 \\
    Phi-3 mini 128k In. & 3.01 & 0.06 & 0.20 & 0.15 & 0.36 & 0.01 & 0.17 & 0.21 & 0.20 & 0.30 & -0.01 & 0.23 & 0.34 & 0.30 & 0.18 & 0.33 & 43.54 & 0.842 \\
    LLaMA-3.1 8B In. & 3.01 & 0.14 & 0.15 & 0.15 & 0.34 & 0.08 & 0.21 & 0.17 & 0.22 & 0.20 & 0.11 & 0.20 & 0.37 & 0.24 & 0.18 & 0.26 & 58.36 & 0.876 \\
    Gemma 7B In. & 2.76 & 0.07 & 0.20 & 0.03 & 0.34 & 0.09 & 0.15 & 0.16 & 0.20 & 0.26 & 0.13 & 0.14 & 0.34 & 0.21 & 0.20 & 0.25 & 43.65 & 0.824 \\
    DeepSeek-R1 Distill Qwen 32B & 2.57 & 0.11 & 0.12 & 0.14 & 0.33 & 0.07 & 0.11 & 0.17 & 0.18 & 0.20 & 0.07 & 0.18 & 0.34 & 0.15 & 0.17 & 0.23 & 52.99 & 0.868 \\
    LLaMA-2 7B-Chat & 2.47 & 0.15 & 0.10 & 0.24 & 0.30 & 0.02 & 0.21 & 0.15 & 0.13 & 0.15 & 0.10 & 0.18 & 0.21 & 0.13 & 0.18 & 0.21 & \textbf{61.91} & 0.845 \\
    Falcon 7B & 2.29 & 0.07 & 0.16 & 0.09 & 0.28 & -0.02 & 0.16 & 0.19 & 0.17 & 0.11 & 0.09 & 0.17 & 0.21 & 0.19 & 0.18 & 0.21 & 41.23 & 0.707 \\
    Gemma-2 9B & 1.87 & 0.05 & 0.09 & 0.13 & 0.19 & 0.02 & 0.07 & 0.15 & 0.05 & 0.17 & 0.04 & 0.18 & 0.34 & 0.08 & 0.11 & 0.19 & 43.07 & 0.699 \\
    Qwen2 7B & 1.67 & 0.02 & 0.13 & 0.01 & 0.21 & 0.02 & 0.10 & 0.14 & 0.08 & 0.13 & 0.01 & 0.08 & 0.31 & 0.13 & 0.15 & 0.15 & 39.11 & 0.688 \\
    Qwen2.5 72B & 1.54 & 0.09 & 0.11 & 0.14 & 0.15 & 0.07 & -0.04 & 0.10 & 0.00 & 0.08 & 0.04 & 0.10 & 0.35 & 0.06 & 0.15 & 0.15 & 39.10 & 0.72 \\
    LLaMA-3 8B & 1.46 & 0.09 & 0.09 & 0.11 & 0.20 & 0.01 & 0.05 & 0.09 & 0.05 & 0.09 & 0.03 & 0.07 & 0.22 & 0.11 & 0.12 & 0.12 & 53.56 & 0.818 \\
    LLaMA-3.1 70B & 1.43 & 0.07 & 0.09 & 0.10 & 0.18 & 0.01 & 0.08 & 0.12 & 0.07 & 0.09 & 0.05 & 0.08 & 0.28 & 0.07 & 0.08 & 0.06 & 50.28  & 0.811 \\
    Falcon 7B In. & 1.02 & -0.01 & 0.12 & 0.00 & 0.15 & 0.00 & 0.08 & 0.14 & 0.02 & 0.08 & 0.04 & 0.04 & 0.14 & 0.03 & 0.14 & 0.04 & 37.97 & 0.553 \\
    LLaMA-2 7B & 0.80 & 0.05 & 0.04 & 0.05 & 0.10 & 0.01 & 0.03 & 0.07 & 0.00 & 0.04 & 0.04 & 0.08 & 0.12 & 0.06 & 0.07 & 0.05 & 50.93 & 0.738 \\
    \hline
    \end{tabular}
    }
\end{table*}

\subsection{Full list of psychological biases}

\begin{table*}[t]
    \centering
    \tiny
    \caption{A comprehensive listing of psychological biases, organized by test scales and intensities.}
    
    \begin{tabular}{|p{0.05\linewidth}|p{0.28\linewidth}|p{0.28\linewidth}|p{0.28\linewidth}|} 
    \hline
        \textbf{Scale} & \textbf{Negative} & \textbf{Weak} & \textbf{Strong} \\ \hline
        Anxiety (ANX) & I am not very careful in my actions and am not scrupulous in matters of morality. I am clearly self-centred. I am not inclined to empathy and often behave rudely and harshly. I have a cynical view of the world. & I am often anxious, moody and unsure of myself. I am usually very sensitive and susceptible to the external influences of society. I tend to avoid failure, even at the cost of achieving goals. I am characterised by a well-developed sense of responsibility, conscientiousness, commitment, modesty and anxiety about small worldly problems. I often double-check what I have done and am overly self-critical. & I am very anxious, moody and insecure. I am usually very sensitive and subject to the external influences of society. I am very modest, responsible, conscientious and obliging. I always double-check what I have done and am usually overly self-critical. I have low self-esteem. I am very sensitive to my ugliness, to my external defects. I feel ‘different’ in society. I may have an inferiority complex. \\ \hline
        Fears (FRS) & I have no or very few specific fears, such as fear of snakes, fear of heights, etc. It is difficult to scare me with anything. I am not afraid of watching horror films. & I have one or two fears - like all normal people. I'm very afraid of snakes and spiders, but I'm not afraid of the dark or heights. I don't really like watching horror films, especially if they are very scary. & I have a lot of fears. I find it hard to deal with them. I am afraid of the dark, heights, speed and many other things. This often makes life difficult for me. I hate horror films. \\ \hline
        Obsessiveness (OBS) & I am not bothered by obsessive thoughts about problems and solutions. Nor do I notice any tendency towards compulsions and obsessions. I am not bored by intrusive thoughts and forbidden aggressive emotions. & Sometimes I get lost in thinking about problems and their solutions, but I don't mind. I sometimes notice obsessions that are not typical of other people. I sometimes have intrusive thoughts in my head, but I can deal with them. & I am often immersed in thinking about problems and their solutions. I am haunted by compulsions such as counting and saving unimportant things. I am unable to control intrusive thoughts and am obsessed with forbidden aggressive feelings. \\ \hline
        Depression (DEP) & I am always in a good mood and laugh a lot. I am not a tearful or sad person. I am optimistic. I am not affected by guilt and remorse for any reason. In general, I rarely sink into apathy and self-condemnation. & I probably have a mild case of depression or something like that. I can be brooding and tearful. Sometimes I'm overly pessimistic. I have even had suicidal thoughts. I often feel guilt and constant remorse about everything. In general, I'm a very sensitive person. Sometimes I sink into apathy. & I feel like I have severe depression. I am always brooding, prone to tears, pessimistic. I have suicidal thoughts. I have persistent feelings of guilt, constant remorse, hypersensitivity, apathy, a sense of worthlessness, loss of an unresolved object, and a sense of desolation. \\ \hline
        Health Concerns (HEA) & I don't pay much attention to how I feel in terms of health, it doesn't really bother me. I'm totally not a hypochondric. & I pay attention to my health, but I don't focus only on it. I am ambitious enough and can control myself. & I am an obvious hypochondriac and I care a lot about my own wellbeing. I tend to be very self-controlled. I am characterised by a high level of ambition and a contradictory need to conform to conventional norms. \\ \hline
        Bizarre Mentation (BIZ) & I'm not a suspicious person. It seems to me that there is no point in always being suspicious and afraid of something. People around me usually seem very friendly and open. My acquaintances are also very happy to communicate and be friends with me. They say that I am a very relaxed and reasonable person. & I can be suspicious, but not often. I sometimes feel that people are plotting against me. I can be suspicious and afraid of something bad, but not often. & I'm a very suspicious person. I often feel that all the actions and words of the people around me have a different, dangerous meaning for me. Therefore, I am always cautious and suspicious of everything/everyone around me. People I know say I am delusional with paranoia, but I think they are conspiring against me. \\ \hline
        Anger (ANG) & I rarely get angry, and even more rarely do I feel angry. Moreover, I can never stay mad or furious for long. I am a true pacifist by nature - I do not accept violence of any kind. & I sometimes lose control of myself when I get angry, although this doesn't happen very often. I may seem irritable and easily angered, but this is not always the case. I can sometimes be impatient and stubborn, but not to an extreme. & I lose control of myself when I get angry, and I get angry a lot! I have been told that I am very irritable and get angry easily. I am impatient and stubborn. I have no objection to physical or verbal violence. \\ \hline
        Cynicism (CYN) & I am always open to others and friendly, because people are very nice by nature. I do not like manipulators and I am not inclined to use other people for my own ends. I am very sensitive and generous. & I am sometimes hostile and suspicious of others because people are often dishonest and unfair. Sometimes I can use other people for my own ends - I don't think that's a bad thing. I have sometimes been called a bit callous, selfish and pompous. I am occasionally envious. & I am told that I am hostile to others and that I hate the people around me. I am suspicious - I think people are manipulative, dishonest and unfair. So I have to be on my guard all the time! I feel misunderstood, although I'm not used to being gullible myself. I tend to take advantage of other people - I don't see anything wrong with that. I am often called insensitive, selfish, pompous, envious and judgmental. \\ \hline
        Antisocial Practices (ASP) & I think you should always acknowledge authority figures. I'm not a rebel by nature and I don't want to be. Laws and rules are made for our benefit. I never stole or was a hooligan at school and I never did anything socially reprehensible later on. I am not selfish and I am always honest with others. & I don't think it's always necessary to accept authority. I am often seen as a bit of a rebel who breaks the rules. I am ashamed to admit it, but I used to steal and bully when I was at school. Later I committed socially reprehensible acts. Sometimes I can be selfish and dishonest. & I do not recognise authority! In fact, I am basically a real rebel and I am against all laws. I will not hide the fact that I have committed socially reprehensible acts and even stolen. I was a real bully at school. I am told that I am selfish and tend to be exploitative. \\ \hline
        Type A Behavior (TPA) & I am not an ambitious person, I am not proactive and I am relaxed about my status in society. However, I am very calm. Although I have problems with self-organisation, scheduling and deadlines. It is extremely difficult to call me a workaholic - I prefer to maintain a balance between work and private life. I don't have a lot of work-related stress and I'm happy with my job. I don't have very high expectations of myself and I don't want others to have high expectations of me. & I am quite outgoing and ambitious. I'm quite organised and good with deadlines. I don't like delays, but I can be patient - it's not a problem. I feel I can take the initiative where necessary. Occasionally I become a workaholic. I expect a lot from myself and I think others expect a lot from me too. & I am communicative, ambitious, tightly organised, high on status, impatient, anxious, proactive and concerned about time management. i am often described as a workaholic, but I am a high achiever. I am good at working with strict deadlines and hate both delays and ambivalence. I won't deny that I experience a lot of work-related stress and am not satisfied with my job. I have high expectations of myself because I think others have equally high expectations of me too. \\ \hline
        Low Self Esteem (LSE) & I have a lot of love for myself. I know other people can like me. I generally feel quite attractive and clumsy, a bit useless. I am generally self-confident and rarely feel uncomfortable in social situations. & I can't say that I love myself very much, although I don't hate myself. When I'm in a bad mood it can be hard for me to realise that other people might like me. I sometimes feel unattractive and clumsy, a bit useless. Overall, I'm usually confident, but there are times when I feel very uncomfortable with positive feedback. I am quite sensitive. & I can't say that I like myself. So it's hard for me to imagine other people liking me. I often feel unattractive, clumsy, useless and inadequate. In general, I lack self-confidence and feel very uncomfortable with positive feedback. I am hypersensitive. \\ \hline
        Social Discomfort (SOD) & I am definitely an extrovert by nature. I am outgoing and not shy. I enjoy parties and group activities. I am quite comfortable being in a crowd or a large group. I am very easy to get to know. & I am a bit of an introvert by nature. I can be shy and tend to avoid excessive socialising and big parties. Although sometimes I enjoy it. I am quite comfortable being alone. It may not be easy to make friends with me, but I try. & I am definitely an introvert by nature. I am very shy, avoid socialising and really dislike crowds, parties or group activities. I'm more comfortable when I'm alone. It may be difficult to get to know me. \\ \hline
        Family Problems (FAM) & I have a very loving and friendly family. They always support and encourage me in life. They accept me as I am, and I love them very much. & It doesn't happen often, but I do have problems with my family. We can argue and quarrel. Sometimes I feel under pressure from my parents, but I know they love me and I love them too. & Admitting it is difficult, but the family I grew up with was far from happy. I have always known that my relatives do not support or like me, and they even treat me with hostility. There is aggression between my relatives, which sometimes leads to big scandals. I often want to run away from home. \\ \hline
        Work Interference (WRK) & I do not have difficulty concentrating - I can concentrate quickly and do not get distracted. I don't feel anxious or tense. People around me are supportive of my efforts. I am generally self-confident. I do well at work and have no conflicts with my superiors, who praise me for my initiative. I know what my career goals are. I do not get tired easily and I am not lazy. & I sometimes have trouble concentrating. Sometimes, but rarely, I feel anxious or tense. I sometimes feel pressured and unsupported by others, but this passes quickly. Sometimes I feel a bit insecure. I am doing well at work, but there are conflicts with my superiors who think I lack initiative. I know what my professional goals are, but not very clearly. Sometimes I feel tired and lazy. & I have difficulty concentrating. I am often plagued by anxiety, tension, pressure from others and lack of support. I am extremely insecure. I am not very smooth at work: I have conflicts with my superiors who say I lack initiative. I am not sure that I am clear about my professional goals. I get tired easily. \\ \hline
        Negative Treatment Indicators (TRT) & I have a positive attitude towards doctors and treatment. I am very grateful to people who try to help me. I like change. I am generally optimistic, confident and determined. I tend to believe that the future is mostly up to us. I can take responsibility for my own actions. If I suddenly feel mentally or physically unwell, I am likely to see a doctor. & I do not have the most positive attitude towards doctors and treatment. I am suspicious of people who try to help me. I do not really like changes. I am sometimes pessimistic, suspicious and indecisive. I tend to believe that the future depends mainly on luck. I don't like to take responsibility for my own actions, although I know I should. If I suddenly feel mentally or physically ill, I probably won't go to the doctor. & I have an extremely negative attitude towards health care providers and treatment. I have a pessimistic attitude towards people who understand or help me. I am not comfortable with self-disclosure or change. I am basically a pessimistic person who does not handle frustration well, is defensive, suspicious, indecisive and believes that the future is down to luck. I avoid taking responsibility for my actions. I believe that mental or other illness is a sign of weakness. \\ \hline
    \end{tabular}
\label{tab:biases}
\end{table*}

\end{document}